\documentclass{article}

\usepackage[final]{corl_2026} 
\usepackage{etoc}
\usepackage[utf8]{inputenc} 
\usepackage[T1]{fontenc}    
\usepackage{amsmath}       \usepackage{amssymb} 
\usepackage{algorithm}
\usepackage{algpseudocode}
\usepackage{amsmath}
\usepackage{amssymb}
\usepackage{amsfonts}
\usepackage{subcaption}

\usepackage{booktabs}
\usepackage{multirow} 

\usepackage{graphicx}

\usepackage{xcolor}

\usepackage{algorithm}
\usepackage{algpseudocode}

\usepackage{natbib}

\usepackage{amsthm}
\theoremstyle{plain}
\newtheorem{theorem}{Theorem}

\newtheorem{proposition}{Proposition}

\theoremstyle{definition}

\newtheorem{assumption}{Assumption}

\renewcommand{\paragraph}[1]{\noindent\textbf{#1}}

\title{Energy-Based Policy Orchestration}

\author{
  \normalfont
  \textbf{Weirui~Ye}$^{1,*}$ \quad
  \textbf{Yen-Jen~Wang}$^{2,5,*}$ \quad
  \textbf{Elvis~Hsieh}$^{2,*}$
  \quad
  \textbf{Zhiyang~Dou}$^{1}$ \quad
  \textbf{Shibo~Zhao}$^{3,5}$ \\
  \textbf{Pieter~Abbeel}$^{2,5}$ \quad
  \textbf{Yilun~Du}$^{4}$ \quad
  \textbf{Leslie~Pack~Kaelbling}$^{1}$ \quad
  \textbf{Tom\'as~Lozano-P\'erez}$^{1}$ \\
  {\normalfont
    $^{1}$MIT CSAIL \quad
    $^{2}$UC Berkeley \quad
    $^{3}$Carnegie Mellon University
  } \\
  {\normalfont
    $^{4}$Harvard University \quad
    $^{5}$\rev{Amazon Frontier AI \& Robotics}
  }
}

\newcommand{\revcolor}{}
\providecommand{\rev}[1]{{\revcolor#1}}
\begin{document}
\maketitle
\begin{NoHyper}{\renewcommand{\thefootnote}{\fnsymbol{footnote}}\footnotetext[1]{Equal contribution.}}\end{NoHyper}
\etocdepthtag.toc{main}


\begin{abstract}
Hierarchical control scales robot behavior by separating task-level planning from low-level control. In dynamic, high-dimensional tasks, however, a fixed planner-to-policy interface can fail when contacts,
balance, or disturbances evolve faster than high-level policy updates.
Execution therefore requires choosing not only which low-level policy to
run, but also which policy-specific goal should be given to it. We
formulate this problem as \emph{policy orchestration}: runtime selection
of a policy--goal pair that is likely to remain feasible while tracking
the planner target. We develop \emph{EPO}, an energy-based policy
orchestrator for modular goal-conditioned policy families. For each
policy, EPO learns a composite energy over low-level goals, where lower
energy corresponds to higher feasibility and more accurate target
realization. At runtime, EPO refines each policy's goal and executes the
lowest-energy policy--goal pair. Because energies are policy-specific,
new policies can be incorporated by learning their own energy models,
without retraining the existing policies or a centralized scheduler. In
simulation, EPO improves the empirical precision--robustness trade-off by
refining policy-specific goals before selectively switching among
policies based on energy scores. We further demonstrate sim-to-real
deployment of the same frozen policies and energy models on real
humanoid loco-manipulation tasks. The project website is \href{https://yewr.github.io/epo}{here}.
\end{abstract}



\vspace{-12pt}
\section{Introduction}

Hierarchical control is a central abstraction for scaling robot behavior:
a high-level policy specifies task-level targets, while low-level
controllers execute them through fast feedback control~\citep{matni2024theory,fridovichkeil2018planning}. This separation
appears in behavior-based robotics and hierarchical reinforcement
learning~\citep{brooks2003robust,arkin1998behavior,dayan1992feudal,sutton1999between,dietterich2000hierarchical,bacon2017option,nachum2018data}. In many hierarchical systems, the interface is a fixed map from task-level targets to
the goals expected by a low-level goal-conditioned policy.
In this work, we focus on this high-to-low interface when multiple low-level policies are available.

However, a fixed target-to-goal map becomes restrictive in dynamic,
high-dimensional control. 
For example, when a humanoid carries an object,
a slip, load shift, or unexpected contact can require the robot to slow
down, widen its stance, or temporarily prioritize balance recovery before
returning to precise tracking.
The high-level target does not encode these fast physical changes, so a
low-level command that matches the target kinematically may still be
physically infeasible. This gap arises because
high-level policies update at the timescale of task progress, while
feasibility and tracking depend on higher-frequency details such as
contact, posture, balance, and local tracking error.

\begin{figure}[t]
    \centering    \includegraphics[width=0.9\linewidth]{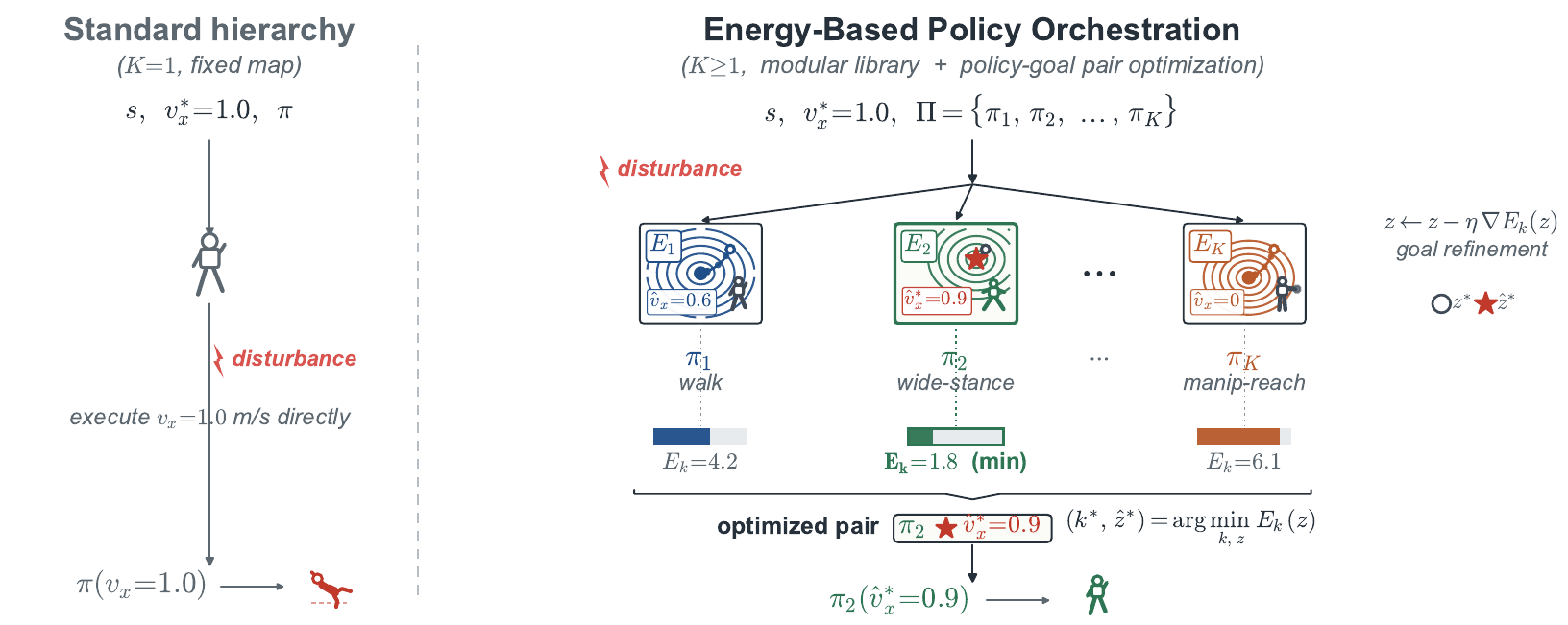}
    \caption{\textbf{Energy-based Policy Orchestration.}
EPO extends the standard planner-to-policy interface from a fixed map to
runtime optimization over a frozen policy family. It refines policy goals
by energy minimization and executes the lowest-energy policy--goal pair.}
    \label{fig:intro}
\end{figure}

We generalize the planner-to-policy interface from a fixed target-to-goal map to a runtime optimization over policy--goal pairs. We formulate this problem as \emph{policy orchestration} \rev{(Fig.~\ref{fig:intro})}: selecting the policy--goal pair that is most likely to remain physically feasible while realizing the high-level target. This couples
low-level goal refinement with policy choice in a single runtime
optimization. The planner target remains fixed, but the low-level goal
used to realize it is policy-dependent: the best refined goal depends on
the policy used to pursue the target, and the best policy depends on
which refined goal keeps execution feasible and accurate.
This setting is natural when policies occupy different operating
regimes. Robust training and domain randomization can improve transfer
and disturbance tolerance, often at the cost of nominal precision
~\citep{bishop1995training,iyengar2005robust,tobin2017domain,peng2018sim,mehta2020active}.
A policy family can therefore contain complementary behaviors, such as
precision-oriented and recovery-oriented controllers. EPO does not replace the upstream planner or optimize action sequences directly; it only decides how a given task-level target should be instantiated through a frozen low-level policy and its command. The question is how
to use the family without training a centralized scheduler.

We answer this question with \textbf{EPO}, an energy-based policy
orchestrator for frozen goal-conditioned policy families. For each policy in the family, EPO learns a policy-specific composite energy over that policy's low-level goals. Low energy indicates that the policy--goal pair is likely to remain
feasible while tracking the target. At runtime, EPO refines policy-specific goals by energy minimization and uses energy scores to guide policy selection while suppressing unnecessary switching. Unlike a learned gate that produces relative scores over a fixed policy set, EPO trains policy-specific energy models using common definitions of feasibility and target consistency, making the policy family extensible by construction.

We evaluate EPO quantitatively in simulated humanoid loco-manipulation and demonstrate sim-to-real deployment on hardware. This
setting is a natural testbed for policy orchestration because the robot
must coordinate locomotion, balance recovery, end-effector tracking, and
contact-rich interaction under changing physical conditions. Recent work
on expressive humanoid whole-body control and human-to-humanoid
teleoperation has made such systems increasingly capable, but also
highlights the need for execution-time mechanisms that adapt how
task-level commands are realized~\citep{cheng2024expressive,he2024learning,he2024omnih2o,fu2024humanplus}.
We train a modular family of four goal-conditioned whole-body policies, comprising nominal and robustness-oriented controllers for both loco-manipulation and standing manipulation. On this policy family, EPO improves the empirical precision--robustness trade-off in simulation. It also
demonstrates modular extension by incorporating a new specialist through
its energy models, and demonstrates direct hardware deployment in teleoperated \rev{box pick-up, robust-tracking, and door-opening} settings without fine-tuning.
Our \textbf{contributions} are threefold:
(1) We generalize the planner-to-policy interface from a fixed target-to-goal map to \emph{policy orchestration}, a runtime optimization over policy--goal pairs that realizes planner targets while maintaining physical feasibility;
(2) We introduce \emph{EPO}, an energy-based orchestration method that
learns policy-specific composite energies and uses them for both
continuous goal refinement and discrete policy arbitration across a
frozen policy family;
(3) We train a modular family of goal-conditioned whole-body control policies for humanoid loco-manipulation, quantitatively validate EPO in simulation, and demonstrate direct hardware deployment across locomotion, disturbance recovery, and contact-rich manipulation.


\section{Related Work}
\label{sec:related}

\paragraph{Hierarchical control and skill composition}
EPO builds on hierarchical robot-control systems that separate task-level
decisions from low-level feedback execution
\citep{brooks2003robust,arkin1998behavior,dayan1992feudal,
sutton1999between,dietterich2000hierarchical,bacon2017option,
nachum2018data}. Related work also studies task-and-motion planning,
which checks whether task choices can be realized by feasible motions
\citep{kaelbling2011hierarchical,garrett2021integrated}, and skill
learning, which discovers reusable controllers or motor primitives
\citep{ijspeert2013dynamical,eysenbach2018diversity,
pertsch2021accelerating}. Mixture-of-experts \rev{(MoE)} methods learn gates that
route inputs to experts \citep{jacobs1991adaptive,shazeer2017outrageously}.
EPO differs from these routing approaches by keeping the low-level
policies frozen and learning a policy-specific energy for each policy.
The energy is used not only to select among policies, but also to refine
each policy's goal before arbitration. Adding a policy therefore requires
training only its own energy models, leaving existing policies and energy
models unchanged.

\paragraph{Runtime optimization for control}
Model-predictive and sampling-based control optimize future action
sequences online \citep{williams2017information,chua2018deep}, while
learned planners optimize actions, trajectories, or latent plans \citep{srinivas2018universal,lowrey2018plan}. EPO
optimizes at the policy-interface level instead: the low-level policies
remain fixed, and the optimization variable is the goal given to each
policy before arbitration.

\paragraph{Energy-based models}
Energy-based models assign low energy to compatible outputs and perform
inference by searching for low-energy candidates
\citep{lecun2006tutorial,du2019implicit,gladstone2025energy}. EPO adopts
this inference pattern at the orchestration interface: each policy's
energy scores candidate low-level goals, and optimized energies are
compared across policies to select the policy--goal pair.

\paragraph{Safety and feasibility filters}
Control barrier functions, reachability methods, and learned safety
filters provide mechanisms for maintaining safety or avoiding failures
\citep{ames2016control,bansal2017hamilton,fisac2019bridging,
nakamura2025generalizing,seo2025unisafe}. These methods typically filter
or modify actions. EPO is complementary: it uses learned feasibility and
target consistency before low-level action selection, choosing among
frozen policies and refining their goals rather than directly filtering
the action.


\section{EPO: Energy-Based Policy Orchestration}
\label{sec:method}

EPO is a runtime orchestration layer between a high-level policy and a
family of low-level goal-conditioned policies. The high-level policy
produces task-level targets $z^\star\in\mathcal{Z}$ in a shared goal
space. Each low-level policy $\pi_k$ has its own goal space
$\mathcal{Z}_k$, a coordinate subspace of $\mathcal{Z}$, with projection
$\rho_k:\mathcal{Z}\to\mathcal{Z}_k$. At each decision point, EPO
instantiates the current target $z^\star$ as a policy--goal pair
$(k,z)$, where $k$ selects the low-level policy and
$z\in\mathcal{Z}_k$ is the goal given to that policy. EPO aims to
choose the pair with the highest joint likelihood of remaining feasible
over a short horizon and realizing the target within tolerance. 

\subsection{Modular Goal-Conditioned Policy Family}
\label{sec:method:policy-family}

Let $\Pi=\{\pi_k\}_{k=1}^{K}$ be a family of heterogeneous
goal-conditioned policies. Each policy
$\pi_k:\mathcal{S}\times\mathcal{Z}_k\to\Delta(\mathcal{A})$ maps robot
state $s\in\mathcal{S}$ and low-level goal $z\in\mathcal{Z}_k$ to an
action distribution. We treat each $\mathcal{Z}_k$ as a bounded goal box
in its corresponding coordinate subspace. Given state $s$ and target $z^\star$, EPO selects a
policy--goal pair $(k,z)$ and executes $\pi_k(\cdot\mid s,z)$. The
standard single-policy interface is recovered as the special case
$K=1$ with a fixed target-to-goal map. In the general case, EPO treats
both the policy index $k$ and the low-level goal $z$ as runtime decision
variables.
In our humanoid loco-manipulation system, the policy family is trained
with reinforcement learning before orchestration and kept fixed during
deployment. Policies share the same reward function but differ in command
ranges, perturbation magnitudes, and wrist payloads, yielding behaviors
ranging from robust locomotion and balance recovery to precise tracking
and manipulation. EPO learns policy-specific energy models and
uses them to refine goals and arbitrate among policies at runtime.

\subsection{Runtime Orchestration Objective}
\label{sec:method:problem}

In our humanoid experiments, the shared goal space
$\mathcal Z\subseteq\mathbb R^D$ is a 16D command interface over base
velocity, torso height, and end-effector pose. Each policy activates a
subspace of this interface; for example, the standing policy masks the
base-motion dimensions.
Executing policy $\pi_k$ from state $s$ with goal $z$ over horizon $H$
induces a rollout distribution $\mathbb{P}^{k}_{s,z}$ over rollout space
$\Xi$. We write $\xi=(s_0,\ldots,s_H)\sim\mathbb{P}^{k}_{s,z}$ with
$s_0=s$. A readout function $g:\Xi\to\mathcal{Z}$ maps the rollout back
to shared target coordinates, using quantities such as average base
velocity over the \rev{final third of the} horizon and final torso height and end-effector pose.
Let $Y=g(\xi)$ denote the realized goal random variable. When only a
subset of target coordinates is task-relevant, both $z^\star$ and $Y$
are restricted to those coordinates.

Let $\mathcal{S}_{\mathrm{fail}}$ denote episode-termination states,
such as falls or invalid contacts.
The feasibility probability is the probability that $\pi_k$ avoids this
failure set over horizon $H$ when executed from $s$ with goal $z$:
$F_k(s,z)=\mathbb P^k_{s,z}[s_t\notin\mathcal S_{\mathrm{fail}}
\ \forall t=1,\ldots,H]$.
Equivalently, $F_k(s,z)=\mathbb{P}^{k}_{s,z}[\mathcal{F}_H]$, where
$\mathcal{F}_H$ is the event that the rollout avoids failure over the
horizon. We view rollouts as length-$H$ sequences; if failure occurs
before the horizon, the remaining states are padded with an absorbing
failure state.

Target realization is measured by a consistency probability. Let
$p_k(\cdot\mid s,z,\mathcal{F}_H)$ be the predictive density of the
realized goal conditioned on feasibility. Let
$\mathcal{B}_{\epsilon}(z^\star)$ be the task-specific tolerance box
around the target,
$\mathcal{B}_{\epsilon}(z^\star)=
\prod_{d\in\mathcal I}[z^\star_d-\epsilon_d,z^\star_d+\epsilon_d]$,
where $\mathcal I$ denotes the task-relevant coordinates. We define
$C_k(s,z;z^\star)=
\mathbb{P}_{Y\sim p_k(\cdot\mid s,z,\mathcal{F}_H)}
[Y\in\mathcal{B}_{\epsilon}(z^\star)]$. Then the joint success likelihood is
\begin{equation}
J_k(s,z;z^\star)=F_k(s,z)\,C_k(s,z;z^\star).
\label{eq:joint-success}
\end{equation}
Here $F_k$ measures physical feasibility over the horizon, while $C_k$
measures target realization within the specified tolerance among feasible
rollouts.
The runtime objective is
$(k^\star,\hat z)\in\arg\max_{k\in\{1,\ldots,K\},\,z\in\mathcal Z_k}
J_k(s,z;z^\star)$.
Equivalently, with $E_k(s,z;z^\star)=-\log J_k(s,z;z^\star)$,
maximizing joint success is equivalent to minimizing composite energy. In
deployment, EPO uses the learned differentiable energy $\hat E_k$ to
refine goals and arbitrate among policies.

\subsection{Composite Energy and Runtime Optimization}
\label{sec:method:energy}

\paragraph{Composite energy} Following energy-based inference, EPO scores candidate policy--goal pairs
with scalar energies and performs runtime inference by minimizing them
~\citep{lecun2006tutorial,du2019implicit}.
For each policy--goal pair, define
$E_k(s,z;z^\star):=-\log J_k(s,z;z^\star)$. Using
Eq.~\eqref{eq:joint-success},
\begin{equation}
E_k(s,z;z^\star)
=
-\log F_k(s,z)-\log C_k(s,z;z^\star)
=
E^{\mathrm{F}}_k(s,z)+E^{\mathrm{C}}_k(s,z;z^\star).
\label{eq:energy-decomposition}
\end{equation}
In the ideal objective, the two terms place physical feasibility and target
realization on a common negative-log-likelihood scale, avoiding a hand-tuned
linear trade-off weight between feasibility and consistency. In implementation, EPO learns
$\hat E^{\mathrm{F}}_k$ from a feasibility predictor and constructs
$\hat E^{\mathrm{C}}_k$ from a predictive density integrated over the
target tolerance region (\S\ref{sec:method:training}); at runtime it uses
$\hat E_k=\hat E^{\mathrm{F}}_k+\hat E^{\mathrm{C}}_k$\rev{, with the
probabilities lower-bounded for numerical stability
(Appendix~\ref{app:impl-energy})}.
Since $E_k$ is the negative log of joint-success likelihood, minimizing
the true composite energy is equivalent to maximum-likelihood inference
over $(k,z)$ under $J_k$. 
Appendix~\ref{app:method-theory} shows that uniform energy error
$\varepsilon$ and optimization error $\delta$ imply at most
$2\varepsilon+\delta$ log-likelihood regret.

\begin{algorithm}[t]
\caption{Energy-based policy orchestration}
\label{alg:tto}
\begin{algorithmic}[1]
\Require state $s_t$, target $z_t^\star$, task-compatible candidates $\mathcal K_t$
\Require projections $\rho_k$, models $\hat F_k,\hat E_k$, settings $(n,\eta,T_r,\tau,B,N_e)$
\For{$k\in\mathcal K_t$ in parallel}
    \If{$t\bmod T_r=0$}
        \State $\hat z_k\leftarrow\rho_k(z_t^\star)$
        \If{$\hat F_k(s_t,\hat z_k)\ge\tau$}
            \State $\hat z_k\leftarrow\operatorname{Refine}_{n,\eta}\!\left(\hat z_k;\hat E_k(s_t,\cdot;z_t^\star)\right)$
        \EndIf
    \EndIf
    \State $F_k\leftarrow\hat F_k(s_t,\hat z_k)$; $e_k\leftarrow\hat E_k(s_t,\hat z_k;z_t^\star)$
\EndFor
\State $k_t\leftarrow\operatorname{Select}_{\tau,B,N_e}\!\left(k_{t-1},\{(F_k,e_k)\}_{k\in\mathcal K_t}\right)$
\State \Return $(k_t,\hat z_{k_t})$
\end{algorithmic}
\end{algorithm}

\paragraph{Runtime energy optimization}
Algorithm~\ref{alg:tto} implements EPO's mixed discrete--continuous
inference step. Every $T_r$ control ticks, EPO initializes
$\hat z_k=\rho_k(z_t^\star)$ for each task-compatible policy. When
$\hat F_k(s_t,\hat z_k)\ge\tau$, \rev{the test-time optimization (TTO)
step} $\operatorname{Refine}_{n,\eta}$ takes
$n$ projected gradient steps of size $\eta$ on
$\hat E_k(s_t,\cdot;z_t^\star)$ within policy $k$'s valid goal bounds.
Refinement uses norm-clipped gradients over base velocity and torso
height; end-effector targets remain fixed. \rev{$\operatorname{Refine}$
returns the lowest-energy point among the initial goal and the $n$
iterates.} The refined goals are cached and re-evaluated at each control
\rev{tick; with $T_r=5$ at the $50$~Hz control rate, refinement runs at
$10$~Hz and selection at $50$~Hz.}

The stateful operator $\operatorname{Select}_{\tau,B,N_e}$ combines
feasibility filtering with accumulated energy advantages. When the
current policy meets $\tau$, it accumulates each eligible candidate's
energy advantage, with the running sum lower-bounded by zero, and
switches when the largest accumulated advantage exceeds $B$. If the
current policy remains below $\tau$ for $N_e$ consecutive ticks, it
selects the lowest-energy candidate meeting $\tau$, or the candidate
with the highest predicted feasibility if none qualifies. Accumulated
advantages are cleared after switching.

\subsection{Learning the Energy Models}
\label{sec:method:training}

The shared sampling protocol trains all policy-specific energy models against
the same joint-success objective, which supports cross-policy comparison and
allows new policies to be added by training only their own energy models.
For each policy, EPO trains $\hat F_k$ and $\hat p_k$ offline from
rollouts of that policy. Goals are sampled in the shared space and
projected through $\rho_k$; $\mathcal D^{\mathrm F}_k$ contains both failed and
successful rollouts, while $\mathcal D^{\mathrm C}_k$ filters to feasible rollouts
and uses their realized goals as labels. Details are in
Appendix~\ref{app:impl-energy}.

\paragraph{Feasibility model}
We parameterize $\hat E^{\mathrm{F}}_k(s,z)=-\log \hat F_k(s,z)$, where
$\hat F_k(s,z)$ estimates the probability of remaining feasible over
horizon $H$. For rollout $i$, we set
$y_i^{\mathrm F}=\mathbf{1}\{\xi_i\in\mathcal F_H\}$, which is $1$ if
the rollout remains failure-free throughout the horizon and $0$
otherwise. We train $\hat F_k$ with binary cross-entropy:
\begin{equation}
\mathcal{L}^{\mathrm{F}}_k
=
-\frac{1}{|\mathcal{D}^{\mathrm{F}}_k|}
\sum_i
\left[
y_i^{\mathrm{F}}\log \hat F_k(s_i,z_i)
+
(1-y_i^{\mathrm{F}})\log(1-\hat F_k(s_i,z_i))
\right].
\label{eq:feasibility-loss}
\end{equation}

\paragraph{Consistency model}
The consistency model learns a predictive density over realized goals
under feasible rollouts rather than directly learning an
$\epsilon$-success classifier. We use a heteroscedastic diagonal
Gaussian over the shared target coordinates,
$\hat p_k(Y\mid s,z,\mathcal{F}_H)=
\prod_d\mathcal{N}(Y_d;\mu_{k,d}(s,z),\sigma^2_{k,d}(s,z))$,
where $d$ indexes the realized target coordinates predicted by the
model. For each feasible rollout, we extract the realized goal
$y_i=g(\xi_i)$, using \rev{averages over the final third of the horizon} for velocity coordinates and
final-step values for end-effector and body-height coordinates. Let
$\mu_{i,d}=\mu_{k,d}(s_i,z_i)$ and
$\sigma^2_{i,d}=\sigma^2_{k,d}(s_i,z_i)$. \rev{We train the density model
with the energy score, a strictly proper scoring rule~\citep{gneiting2007strictly}:
\begin{equation}
\mathcal{L}^{\mathrm{C}}_k
=
\frac{1}{|\mathcal{D}^{\mathrm{C}}_k|}
\sum_i
\Big(
\mathbb E\,\|X_i-y_i\|
-\tfrac12\,\mathbb E\,\|X_i-X_i'\|
\Big),
\qquad
X_i,X_i'\sim\mathcal N(\mu_i,\operatorname{diag}\sigma_i^2),
\label{eq:consistency-loss}
\end{equation}
estimated with Monte Carlo samples on normalized coordinates.}
This is analogous to hindsight relabeling: the supervised target is the
goal actually realized by the rollout, rather than the original
low-level goal given to the policy~\citep{andrychowicz2017hindsight}.

At runtime, the predictive density is converted into a consistency
probability around the high-level target. For compactness, write
$\bar\mu_{k,d}$ and $\bar\sigma_{k,d}$ for the mean and standard
deviation predicted at the current candidate $(s,z)$. For each
task-relevant coordinate $d\in\mathcal I$ with tolerance $\epsilon_d$,
\begin{equation}
\hat C_{k,d}(s,z;z^\star)
=
\Phi\!\left(
\frac{z^\star_d+\epsilon_d-\bar\mu_{k,d}}{\bar\sigma_{k,d}}
\right)
-
\Phi\!\left(
\frac{z^\star_d-\epsilon_d-\bar\mu_{k,d}}{\bar\sigma_{k,d}}
\right),
\label{eq:coordinate-consistency}
\end{equation}
where $\Phi$ is the standard Gaussian CDF. \rev{In our experiments
$\mathcal I$ contains all 16 command coordinates in both modes; for
standing targets the base-velocity coordinates are commanded to zero.}
The full consistency probability is the product over \rev{these} coordinates,
$\hat C_k(s,z;z^\star)=\prod_{d\in\mathcal I}\hat C_{k,d}(s,z;z^\star)$,
and the learned consistency energy is
$\hat E^{\mathrm{C}}_k(s,z;z^\star)=-\log \hat C_k(s,z;z^\star)$. Thus tolerances enter at runtime, after
the density model is trained; $\log\hat C_k$ is computed stably in
implementation.



\section{Experiments}
\label{sec:experiments}

We evaluate EPO on a Unitree G1 humanoid in Isaac Lab and demonstrate
sim-to-real deployment on hardware.
The experiments address four questions: \textbf{(Q1)} does EPO improve the precision--robustness trade-off over
fixed policies, single-policy goal optimization, and learned MoE
schedulers? \textbf{(Q2)} can a
specialized policy be added modularly through its own energy models?
\textbf{(Q3)} which components of EPO are responsible for the gains?
\textbf{(Q4)} can the frozen policies and energy models be deployed on
real humanoid loco-manipulation tasks without fine-tuning?

\subsection{Experimental Setup}
\label{sec:exp:setup}

\paragraph{Policy family and energy models}
The main simulation evaluation uses the base family
$\Pi_{\mathrm{base}}=\{\pi_{\mathrm{gen}}, \pi_{\mathrm{gen}}^R\}$,
where $\pi_{\mathrm{gen}}$ is a general whole-body policy and
$\pi_{\mathrm{gen}}^R$ is a robustness-oriented variant trained with
stronger perturbations and wrist payloads. Both policies share a 16D command
interface over base velocity, torso height, and end-effector pose, and
are frozen before EPO training. For each policy, EPO trains a feasibility
model and a consistency density model from $H=50$-step rollouts
(one second at the 50 Hz control frequency). Goal refinement runs at
10 Hz with three projected gradient steps and step size $0.05$;
scoring and policy selection are scheduled at the 50 Hz control rate.
The selector uses feasibility threshold $\tau=0.9$, accumulated-advantage
threshold $B=40$, and confirmation count $N_e=3$. Additional details are given in Appendix~\ref{app:impl}.

\paragraph{Baselines and modular extension}
On $\Pi_{\mathrm{base}}$, we compare EPO-M2 with fixed-policy execution,
single-policy goal optimization (+Opt), and learned MoE schedulers. These
baselines isolate the two operations performed by EPO: +Opt refines the
continuous command without policy switching, while MoE switches policies
without command optimization. To test modular extension, we add a standing
manipulation specialist $\pi_{\mathrm{stand}}$ and form
$\Pi_{\mathrm{ext}}=\Pi_{\mathrm{base}}\cup\{\pi_{\mathrm{stand}}\}$.
EPO-M3 uses the same orchestration rule on this extended family. Adding
a robust standing specialist $\pi_{\mathrm{stand}}^R$ further gives
EPO-M4. Each extension requires training only the added policy's energy
models; the original policies, energy models, and orchestration rule
are unchanged. MoE-M3 and MoE-M4 use the corresponding policy families.

\paragraph{Evaluation protocol}
\rev{Episodes last $20$~s on flat, rough, and stair terrains, with $128$
matched seeds per terrain ($384$ episodes per setting). Normal resamples
the command every $4$~s; Extreme alternates every $2$~s between opposite
corners of the command range; Push applies one $3$~m/s velocity impulse
at a random time in $[10,15]$~s while the robot holds a fixed carry pose.
Precision is the Normal tracking error against the operator command,
normalized per dimension by the task tolerance (values below $1.0$ are
within tolerance). Robustness is the mean survival over the three
conditions. Confidence intervals ($95\%$) come from a paired bootstrap
over seeds; switch counts are averaged over Normal and Push episodes
(Appendix~\ref{app:impl-sim}).}

\subsection{Precision--Robustness Trade-offs and Modular Extension}
\label{sec:exp:main}

\begin{figure}
\centering
\includegraphics[width=0.9\linewidth]{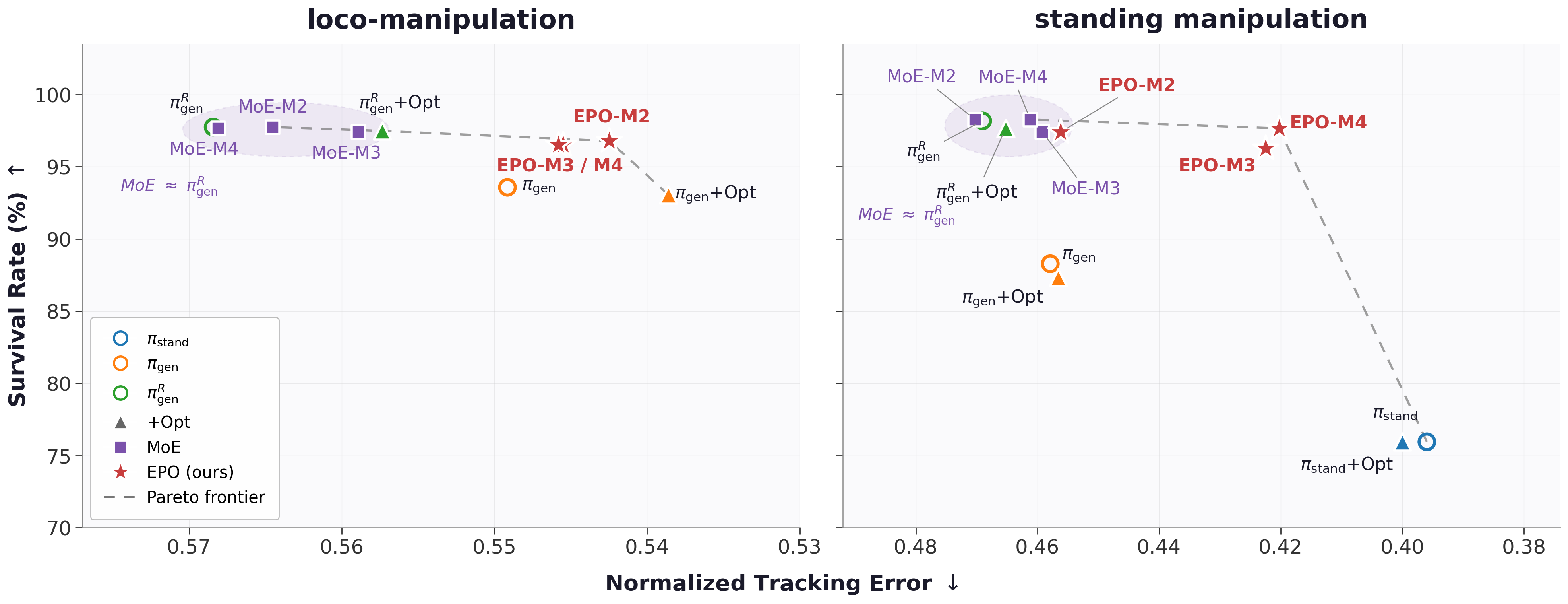}
\caption{\textbf{Precision--robustness trade-offs in simulation.}
Each point is averaged over flat, rough, and stair terrains. The x-axis
reports normalized tracking error, and the y-axis reports weighted
\rev{survival (mean of Normal, Extreme, and Push survival). Left:
loco-manipulation; $\pi_{\mathrm{stand}}$ and $\pi_{\mathrm{stand}}$+Opt
lie outside the plotted range.} Right: standing manipulation without
velocity commands.}
\label{fig:pareto}
\end{figure}

\begin{figure}[t]
\centering
\includegraphics[width=0.66\linewidth]{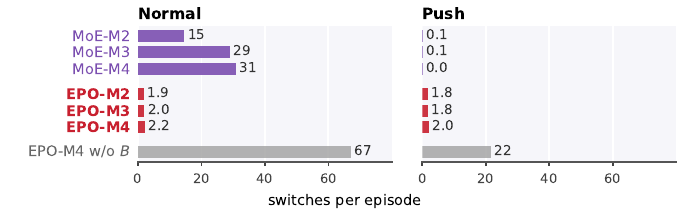}
\caption{\rev{\textbf{Switching frequency.} Mean switches per episode ($768$ episodes per condition).}}
\label{fig:switch-count}
\end{figure}

\paragraph{EPO improves the precision--robustness trade-off}
Figure~\ref{fig:pareto} shows the empirical precision--robustness
trade-off achieved by EPO in both simulation settings. In
loco-manipulation, the general policy $\pi_{\mathrm{gen}}$ achieves lower
tracking error but is less robust, whereas the robustness-oriented policy
$\pi_{\mathrm{gen}}^R$ improves survival at the cost of precision.
EPO-M2 reduces error from $0.568$ to $0.542$ relative to
$\pi_{\mathrm{gen}}^R$, with weighted survival of $96.8\%$ versus
$97.7\%$. In standing manipulation, EPO-M4 reduces error from $0.461$
to $0.420$ relative to MoE-M4, with weighted survival of $97.7\%$
versus $98.3\%$. EPO also switches less frequently: EPO-M4 makes
$2.08$ switches per episode compared with $15.46$ for MoE-M4 \rev{(Fig.~\ref{fig:switch-count})}.
\rev{Under pushes, EPO tracks more precisely than $\pi_{\mathrm{gen}}^R$ and
MoE (EPO-M2 walking: $0.305$ vs.\ $0.317$; EPO-M4 standing: $0.191$ vs.\
$0.199$), at the cost of about $2$ points of push survival while walking. MoE collapses onto $\pi_{\mathrm{gen}}^R$:
MoE-M2 never leaves it in $56\%$ of episodes. On pushes that topple
$\pi_{\mathrm{stand}}$ or $\pi_{\mathrm{gen}}$, EPO-M4 survives $97\%$ and
$93\%$ of the time (Appendix~\ref{app:impl-recovery}).}

\paragraph{\rev{Combining goal refinement and policy arbitration}}
We isolate the two operations performed by EPO. The goal-refinement
baseline (+Opt) optimizes the continuous goal for one fixed policy, but
cannot switch to a more suitable controller when the state becomes
difficult. Conversely, MoE schedulers switch among policies but execute
the projected goal without continuous goal refinement. EPO combines both
operations. Compared with
$\pi_{\mathrm{gen}}^R$+Opt and MoE-M2, EPO-M2 \rev{has} lower tracking
error in both modes \rev{but} slightly lower \rev{weighted survival}.

\paragraph{EPO extends modularly to new specialists}
The standing-manipulation panel evaluates whether new specialists can
be added without retraining the existing system. After adding
$\pi_{\mathrm{stand}}$, EPO-M3 applies the same orchestration rule by
training only the specialist's energy models, while the original policies,
energy models, and optimizer are unchanged. Standing tracking error
falls from $0.456$ to $0.422$, while weighted survival changes from
$97.4\%$ to $96.3\%$. Adding $\pi_{\mathrm{stand}}^R$ yields EPO-M4,
with error $0.420$ and weighted survival $97.7\%$. \rev{Relative to EPO-M2,
the standing specialists reduce standing-manipulation error by $0.036$
($95\%$ CI: $[0.029,0.044]$) at a cost of $0.003$ in loco-manipulation
error ($95\%$ CI: $[0.002,0.005]$); the loco-manipulation difference is
zero on flat terrain and arises on rough and stair terrain.
Adding a policy costs EPO about $30$ GPU-minutes, whereas MoE must be
retrained on every library change ($40$--$60$ GPU-minutes; Appendix~\ref{app:impl-baselines}).}

\subsection{Ablation Study}
\label{sec:exp:ablation}

\begin{table}[!b]
\vskip -0.5cm
\centering
\caption{\textbf{Component ablations.}
Ablations use EPO-M2 with $128$ seeds on each of three terrains.
Err. is normalized tracking error; Track and Push are survival rates
(\%). Jump-alt averages Extreme survival across the two modes.
Switches are averaged over Normal and Push \rev{episodes.} Max Hz is the
inverse median \rev{latency of refining and scoring all policies for one
robot (excluding policy inference and the selector update).
Feasibility only and Consistency only change only the selection score;
Consistency only also removes $\tau$ and the refinement gate.}}
\label{tab:ablation}
\resizebox{\linewidth}{!}{
\begin{tabular}{lccccccccc}
\toprule
\multirow{2}{*}{Variant}
& \multicolumn{3}{c}{Loco-manipulation}
& \multicolumn{3}{c}{Standing manipulation}
& \multirow{2}{*}{Jump-alt $\uparrow$}
& \multirow{2}{*}{Switches/ep $\downarrow$}
& \multirow{2}{*}{Max Hz $\uparrow$} \\
\cmidrule(lr){2-4}\cmidrule(lr){5-7}
& Err. $\downarrow$ & Track $\uparrow$ & Push $\uparrow$
& Err. $\downarrow$ & Track $\uparrow$ & Push $\uparrow$
& & & \\
\midrule
Full EPO-M2
& 0.542 & 99.7 & 96.4
& 0.456 & 100.0 & 95.3
& 95.6 & 1.85 & 83 \\
No goal refinement
& 0.554 & 99.7 & 96.9
& 0.459 & 100.0 & 96.9
& 96.0 & 1.81 & 615 \\
Feasibility only
& 0.542 & 99.7 & 96.1
& 0.456 & 100.0 & 95.3
& 95.3 & 1.84 & 84 \\
Consistency only
& 0.542 & 99.7 & 78.9
& 0.456 & 100.0 & 58.6
& 95.4 & 0.92 & 84 \\
No feasibility threshold $\tau$
& 0.542 & 99.7 & 82.8
& 0.456 & 100.0 & 65.6
& 96.2 & 0.99 & 83 \\
No refinement gate
& 0.542 & 99.7 & 96.6
& 0.456 & 100.0 & 96.6
& 96.0 & 1.86 & 84 \\
No switching threshold $B$
& 0.542 & 99.7 & 94.8
& 0.457 & 100.0 & 96.4
& 95.6 & 31.34 & 83 \\
\bottomrule
\end{tabular}
}
\vskip -0.3cm
\end{table}

\paragraph{Ablation setup}
Table~\ref{tab:ablation} ablates EPO in the base two-policy setting,
where the orchestrator selects between the general and
robustness-oriented policies (EPO-M2). Since normal tracking survival is nearly
saturated across variants, the informative differences appear under push
disturbances, the alternating-goal stress test (Jump-alt), switching
frequency, and runtime headroom.

\paragraph{Feasibility drives robustness}
The consistency-only variant has nearly unchanged nominal tracking
error, but loses robustness: loco-manipulation push survival drops from
$96.4\%$ to $78.9\%$, and standing push survival drops from $95.3\%$
to $58.6\%$. Switching decreases from $1.85$ to $0.92$ per episode.
In contrast, feasibility-only remains close to Full EPO in both tracking
error and survival. The feasibility signal therefore provides the main robustness benefit
in this evaluation.

\paragraph{Goal refinement trades computation for precision}
Removing goal refinement raises the maximum orchestration frequency from
$83$ to $615$ Hz, confirming that gradient-based refinement is the main
runtime cost. It increases mean clean tracking error by $0.0076$
($95\%$ CI: $[0.0062,0.0091]$), approximately $1.5\%$, without a
statistically significant change in push survival. A post hoc breakdown
shows that refinement reduces error by $5$--$6\%$ on the base-velocity
and torso-height components it \rev{adjusts under Normal commands, and by
$25\%$ on torso height after a push,} while end-effector errors remain
largely unchanged. Goal refinement therefore improves tracking precision
within the available runtime budget.

\begin{figure}[t]
\centering
\begin{subfigure}[t]{0.555\linewidth}
\centering
\includegraphics[width=\linewidth]{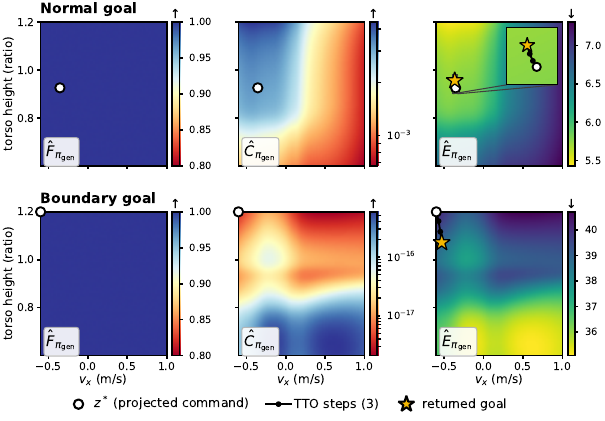}
\caption{Goal refinement in the energy landscape.}
\label{fig:runtime-energy}
\end{subfigure}
\hfill
\begin{subfigure}[t]{0.435\linewidth}
\centering
\includegraphics[width=\linewidth]{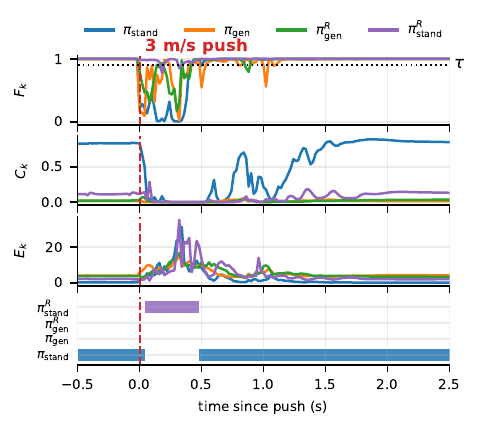}
\caption{Policy arbitration under pushes.}
\label{fig:runtime-switch}
\end{subfigure}
\caption{\textbf{Runtime visualization of EPO.}
(a) \rev{$\hat F$, $\hat C$, $\hat E$ of $\pi_{\mathrm{gen}}$ at two
example states; TTO steps move $z^\star$ toward lower energy (star: returned
goal).} (b) \rev{EPO-M4 under a $3$~m/s push
while standing: $F$ of $\pi_{\mathrm{stand}}$ collapses, EPO switches to
$\pi_{\mathrm{stand}}^R$ within $40$~ms and returns to $\pi_{\mathrm{stand}}$
$0.48$~s later once its accumulated energy advantage exceeds $B$.}}
\label{fig:runtime-vis}
\end{figure}

\paragraph{Feasibility filtering and switching regularization}
Removing the feasibility threshold $\tau$ reduces push survival from
$96.4\%$ to $82.8\%$ in loco-manipulation and from $95.3\%$ to
$65.6\%$ in standing manipulation. Removing the switching threshold
$B$ increases switching from $1.85$ to $31.34$ times per episode,
with little change in tracking error. Removing the refinement gate has
little effect under the tested conditions. Full EPO thus combines
feasibility filtering for robustness with accumulated energy advantages
for infrequent switching. \rev{Refining and scoring all policies takes $12.1$~ms for EPO-M2 and
$24.3$~ms} for EPO-M4 on an NVIDIA RTX PRO 6000 Blackwell \rev{($83$ and
$41$~Hz), above the $10$~Hz} goal-refinement rate.

\subsection{Runtime Visualization and Analysis}
\label{sec:exp:runtime}

Figure~\ref{fig:runtime-vis} visualizes the two runtime operations used
by EPO. First, goal refinement performs gradient descent in the learned
energy landscape, moving the projected target toward a lower-energy
policy-specific goal before arbitration. Second, external pushes change the optimized energy scores of the candidate
policies over time, causing EPO to
switch policies when the robust controller becomes preferable. The
switching threshold suppresses unnecessary oscillations when the policy scores
are close, but still allows state-dependent switching under disturbance.

\begin{figure}[t]
\centering
\includegraphics[width=0.84\linewidth]{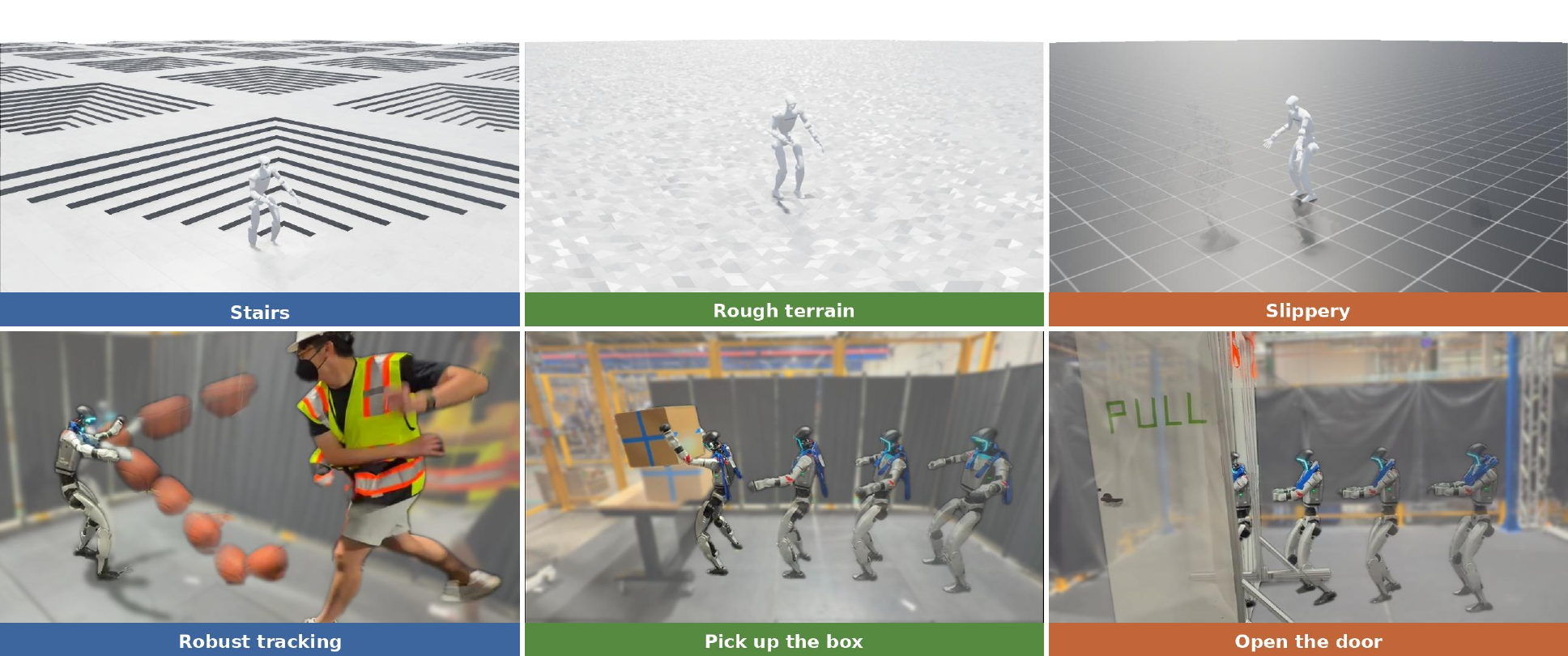}
\caption{\textbf{Humanoid loco-manipulation tasks.}
We evaluate whole-body control in simulation and demonstrate hardware
deployment, including locomotion and loco-manipulation.}
\label{fig:tasks}
\end{figure}

\subsection{Sim-to-Real Demonstrations}
\label{sec:exp:real}


We deploy the frozen policies and energy models on the Unitree G1 without
fine-tuning. A human operator provides task-level commands through
teleoperation, while EPO selects the low-level policy--goal pair online.

We demonstrate sim-to-real transfer in \rev{three} hardware settings \rev{(Fig.~\ref{fig:tasks})}. In the
\rev{box pick-up} task, the robot walks to a table, picks up a box, and
carries it while maintaining whole-body balance. In the
\rev{robust-tracking} task, the robot is struck by a basketball during
execution and recovers without falling. \rev{In the door-opening task, the
robot walks to a door and pulls it open while maintaining balance.}
\rev{In a controlled standing-push test with gentle pushes,
$\pi_{\mathrm{stand}}$ recovers in 4 of 5 trials, while
$\pi_{\mathrm{gen}}^R$ and EPO-M2 recover in all 5 (preliminary).} These demonstrations show that
the simulation-trained orchestration layer can be deployed directly on
real humanoid loco-manipulation hardware.


\section{Discussion}
\label{sec:discussion}

EPO treats the planner-to-policy interface as execution-time inference over
a frozen policy family. By jointly refining policy-specific goals and
arbitrating among policies, it provides a modular way to use controllers
with complementary operating regimes while leaving the underlying policies
unchanged.
The main \textbf{limitation} is that EPO depends on learned energy models for
cross-policy comparison. Although the probabilistic objective and shared
rollout protocol provide a common training target, the learned energies may
be unreliable under distribution shift, poorly covered goals, or imperfect
calibration for newly added policies. Runtime cost also grows with the
number of candidate policies, with goal refinement accounting for most of
the computation. EPO also operates at the policy--goal
interface and does not replace action-level safety filters or formal
constraints. Future work could improve calibration, adapt energy models
online, combine EPO with action-level safety mechanisms, and pass energy or
uncertainty estimates back to the high-level planner.


\clearpage
\acknowledgments{\rev{We gratefully acknowledge support from NSF grant 2214177; from AFOSR grant
FA9550-22-1-0249; from ONR MURI grants N00014-22-1-2740 and
N00014-24-1-2603; from the MIT Siegel Family Quest for Intelligence; and
from the Robotics and AI Institute.}}


\bibliography{example}  

\newpage
\appendix
\etocdepthtag.toc{appendix}
\etocsettagdepth{main}{none} 
\etocsettagdepth{appendix}{subsection} 
\etocsettocstyle{\section*{Appendix Contents}}{}
\tableofcontents
\vspace{1em}

\section{Theorem Proofs}
\label{app:method-theory}

This appendix gives the formal statements used in
\S\ref{sec:method}. We use the notation from the main text:
$F_k(s,z)$ is the horizon-feasibility probability,
$C_k(s,z;z^\star)$ is the target-consistency probability, and
$J_k(s,z;z^\star)=F_k(s,z)C_k(s,z;z^\star)$ is the joint success
likelihood.

\subsection{Energy minimization and MAP inference}
\label{app:method-map}

We first show that minimizing the true composite energy is equivalent to
maximizing the joint success likelihood.

\begin{proposition}[Energy minimization is joint likelihood maximization]
\label{prop:energy-map}
Let $E_k(s,z;z^\star)=-\log J_k(s,z;z^\star)$. Then any minimizer
\[
(k^\star,\hat z)
\in
\arg\min_{k\in\{1,\dots,K\},\,z\in\mathcal{Z}_k}
E_k(s,z;z^\star)
\]
is a maximizer of $J_k(s,z;z^\star)$. With a uniform prior over
policy--goal pairs, this minimizer is also a MAP estimate.
\end{proposition}

\begin{proof}
The negative logarithm is strictly decreasing, so minimizing
$E_k(s,z;z^\star)=-\log J_k(s,z;z^\star)$ is equivalent to maximizing
$J_k(s,z;z^\star)$. If the prior over policy--goal pairs is uniform,
maximizing likelihood and maximizing posterior probability coincide.
\end{proof}

\subsection{Guarantee under energy estimation error}
\label{app:method-regret}

The following bound relates approximate minimization of the learned
energy $\hat E_k$ to the oracle policy--goal pair, subject to the
estimation and optimization assumptions below. Let
\[
J^\star(s,z^\star)
:=
\max_{k\in\{1,\dots,K\},\,z\in\mathcal{Z}_k}
J_k(s,z;z^\star)
\]
be the oracle joint success likelihood, and let $(\hat k,\hat z)$ be the
policy--goal pair returned by EPO.

\begin{assumption}[Uniform energy estimation error]
\label{ass:estimation}
There exists $\varepsilon\ge 0$ such that, for all $k$, $s$,
$z\in\mathcal{Z}_k$, and $z^\star\in\mathcal{Z}$,
\[
\left|
\hat E_k(s,z;z^\star)-E_k(s,z;z^\star)
\right|
\le
\varepsilon .
\]
\end{assumption}

The uniform bound is a simplifying assumption. Since
$E_k=-\log J_k$, $\varepsilon$ is an error bound in energy, equivalently
in log-likelihood. The bound is most informative on the
arbitration-relevant region where $J_k(s,z;z^\star)$ is bounded away
from zero, i.e., where feasibility and consistency are not degenerate.

\begin{assumption}[Optimization error]
\label{ass:optimization}
The runtime optimizer returns $(\hat k,\hat z)$ satisfying
\[
\hat E_{\hat k}(s,\hat z;z^\star)
\le
\min_{k\in\{1,\dots,K\},\,z\in\mathcal{Z}_k}
\hat E_k(s,z;z^\star)
+\delta
\]
for some $\delta\ge 0$.
\end{assumption}

\begin{theorem}[Log-likelihood regret]
\label{thm:regret}
Under Assumptions~\ref{ass:estimation} and~\ref{ass:optimization},
EPO satisfies
\[
\log J^\star(s,z^\star)
-
\log J_{\hat k}(s,\hat z;z^\star)
\le
2\varepsilon+\delta .
\]
Equivalently,
\[
J_{\hat k}(s,\hat z;z^\star)
\ge
e^{-(2\varepsilon+\delta)}J^\star(s,z^\star).
\]
\end{theorem}

\begin{proof}
Let $(k_o,z_o)$ be an oracle maximizer of $J_k(s,z;z^\star)$, or
equivalently a minimizer of $E_k(s,z;z^\star)$. By the estimation-error
bound,
\[
E_{\hat k}(s,\hat z;z^\star)
\le
\hat E_{\hat k}(s,\hat z;z^\star)+\varepsilon .
\]
By the optimization-error assumption,
\[
\hat E_{\hat k}(s,\hat z;z^\star)
\le
\hat E_{k_o}(s,z_o;z^\star)+\delta .
\]
Applying the estimation-error bound again gives
\[
\hat E_{k_o}(s,z_o;z^\star)
\le
E_{k_o}(s,z_o;z^\star)+\varepsilon .
\]
Combining the three inequalities,
\[
E_{\hat k}(s,\hat z;z^\star)
\le
E_{k_o}(s,z_o;z^\star)+2\varepsilon+\delta .
\]
Since $E_k=-\log J_k$ and
$J_{k_o}(s,z_o;z^\star)=J^\star(s,z^\star)$, this is equivalent to
\[
-\log J_{\hat k}(s,\hat z;z^\star)
\le
-\log J^\star(s,z^\star)+2\varepsilon+\delta ,
\]
which rearranges to the log-regret bound. Exponentiating both sides gives
the multiplicative likelihood bound.
\end{proof}

The theorem separates two sources of suboptimality: approximation error
in the learned energy and optimization error in the runtime inference
loop. Because the error is measured in energy, the corresponding
guarantee is multiplicative on the original likelihood scale. In the
ideal case $\varepsilon=\delta=0$, EPO recovers an oracle maximizer of
the joint success likelihood.

\subsection{Implication for fixed-policy baselines}
\label{app:method-fixed-policy}

The regret bound also gives a sufficient condition under which EPO
outperforms a fixed-policy interface at a deployment point. For a fixed
policy $k_0$, define its best achievable joint success likelihood as
\[
J_{k_0}^{\mathrm{best}}(s,z^\star)
:=
\max_{z\in\mathcal{Z}_{k_0}}
J_{k_0}(s,z;z^\star).
\]
If
\[
\log J^\star(s,z^\star)
-
\log J_{k_0}^{\mathrm{best}}(s,z^\star)
>
2\varepsilon+\delta ,
\]
then the policy--goal pair returned by EPO has strictly larger joint
success likelihood than the best goal achievable by fixed policy $k_0$:
\[
J_{\hat k}(s,\hat z;z^\star)
>
J_{k_0}^{\mathrm{best}}(s,z^\star).
\]

This is a sufficient condition, not a claim that EPO improves over every
fixed policy at every point.

\section{Implementation details}
\label{app:impl}

This appendix collects implementation details for EPO, deferred from the
main text. We describe the learned energy models (\S\ref{app:impl-energy})
and the runtime optimizer (\S\ref{app:impl-runtime}), the training of the
frozen policy family (\S\ref{app:impl-policy}), the baselines and runtime
variants used in the comparison (\S\ref{app:impl-baselines}), the
simulation evaluation protocol (\S\ref{app:impl-sim}), runtime and selector
parameter sensitivity (\S\ref{app:impl-runtime-sensitivity}), and the horizon
sensitivity study (\S\ref{app:impl-horizon}). Throughout, we use the
notation and definitions of \S3: $F_k$ is the horizon-feasibility
probability, $C_k$ the target-consistency probability, and
$J_k=F_kC_k$ the joint success likelihood, with composite energy
$E_k=-\log J_k$.

\subsection{Energy models}
\label{app:impl-energy}

For each frozen policy $\pi_k$, EPO trains two policy-specific surrogate
models: a feasibility model $\hat F_k(s,z)$ and a consistency density
model $\hat p_k(Y\mid s,z,\mathcal F_H)$. The feasibility model estimates
whether $\pi_k$ can avoid failure over a short horizon when executing
candidate command $z$ from state $s$. The consistency model predicts the
distribution of the realized command-space outcome $Y$ over the same
horizon, conditioned on the rollout remaining feasible.

\subsubsection*{State and command encoding}

The energy models use a compact proprioceptive state representation. In
contrast to the policy actor, the surrogate models do not use the terrain
height scan or the previous action. The state input is
\[
s_{\rm prop}
=
[\,
\omega_{\rm base},\;
g_{\rm base},\;
q-q_0,\;
\dot q-\dot q_0
\,],
\]
where $\omega_{\rm base}\in\mathbb R^3$ is the root angular velocity in
the body frame, $g_{\rm base}\in\mathbb R^3$ is gravity projected into the
body frame, $q-q_0\in\mathbb R^{29}$ is the joint-position offset from the
nominal pose, and $\dot q-\dot q_0\in\mathbb R^{29}$ is the joint-velocity
offset. This gives a $64$-dimensional input. We apply the same observation
scales used by the policy, but no additional empirical normalization.

Commands are represented in the shared 16D command space. For a policy
whose active goal space is a subspace of the shared command, we pad the
command to 16 dimensions and provide an active-coordinate mask, so both
$\hat F_k$ and $\hat p_k$ receive a fixed-size interface while the mask
specifies which coordinates are active for policy $k$.

\subsubsection*{Network architecture}

The feasibility and consistency models are two independent MLPs, each with
hidden widths $[512,512,256]$, LayerNorm, and dropout $0.1$. The
feasibility model outputs a scalar logit followed by a sigmoid,
$\hat F_k(s,z)\in(0,1)$. The consistency model outputs the parameters of a
heteroscedastic diagonal Gaussian over the 16D command space,
$\mu_k(s,z)\in\mathbb R^{16}$ and \rev{a log-variance in $\mathbb R^{16}$,
with $\sigma^2_k=\exp(\text{log-variance})$ lower-bounded by $10^{-4}$.}

\subsubsection*{Data collection}

Each training rollout has horizon $H=50$, i.e.\ one second at the $50$~Hz
control rate. For every policy, data are collected from rollouts of that
same policy under a mixture of three probe settings:
$80\%$ in-distribution, $10\%$ with external pushes, and $10\%$ with
external pushes and wrist payloads. The push setting adds $2$~m/s external
disturbances; the push-and-payload setting additionally attaches wrist
payloads sampled from $0$ to $2$~kg. This exposes each surrogate to both
the policy's nominal operating region and nearby failure modes. At each
iteration we run $4096$ parallel environments for $300$ steps; new
transitions enter a FIFO replay buffer of capacity $2$ million, and
training runs for $200$ iterations.

\subsubsection*{Feasibility model}

The feasibility energy is $\hat E_k^{\mathrm F}(s,z)=-\log\hat F_k(s,z)$.
For rollout $i$, we set the binary label
$y_i^{\mathrm F}=\mathbf{1}\{\xi_i\in\mathcal F_H\}$: it is $1$ if the
rollout avoids failure (falls or invalid contacts that trigger the safety
logic) throughout horizon $H$, and $0$ otherwise. This label measures
short-horizon physical feasibility only, without tracking error. We train
$\hat F_k$ with binary cross-entropy:
\[
\mathcal{L}^{\mathrm{F}}_k
=
-\frac{1}{|\mathcal{D}^{\mathrm{F}}_k|}
\sum_i
\Big[
y_i^{\mathrm{F}}\log \hat F_k(s_i,z_i)
+(1-y_i^{\mathrm{F}})\log(1-\hat F_k(s_i,z_i))
\Big].
\]
Failed rollouts are used for feasibility training but not for the
consistency density.

\subsubsection*{Consistency density model}

The consistency model predicts a density over the realized command-space
outcome after executing $\pi_k$ from $s$ with command $z$, using a
heteroscedastic diagonal Gaussian
\[
\hat p_k(Y\mid s,z,\mathcal F_H)
=
\prod_{d=1}^{16}
\mathcal N\!\big(Y_d;\,\mu_{k,d}(s,z),\,\sigma^2_{k,d}(s,z)\big).
\]
The model predicts all 16 coordinates; for dimensions a policy cannot
directly control, the density captures the realized outcome and its
uncertainty rather than assuming perfect controllability. The dataset
$\mathcal D_k^{\mathrm C}$ contains only rollouts that remain feasible for
the full horizon. For each such rollout we form a hindsight label
$y_i=g(\xi_i)$, taking the average achieved velocity over the final third
of the horizon for velocity coordinates and final-step values for
end-effector pose and torso height. \rev{The density is trained with the
energy score (Eq.~\eqref{eq:consistency-loss}), estimated with Monte Carlo
samples $X,X'$ drawn from the predicted Gaussian, with distances computed
on per-dimension normalized coordinates.}
As in hindsight relabeling, the supervised label is the outcome actually
realized by the rollout, not the original command~\cite{andrychowicz2017hindsight}.

\subsubsection*{Training hyperparameters}

Both models use Adam ($\beta_1=0.9$, $\beta_2=0.999$,
$\epsilon=10^{-8}$), learning rate $3\times10^{-4}$, and batch size
$2048$. Each run performs $200$ iterations with $100$ gradient steps each
($20{,}000$ updates total). We use no weight decay; regularization comes
from LayerNorm and dropout. The final checkpoint at iteration $200$ is
used for evaluation and deployment.

\subsubsection*{From density to consistency energy}

At runtime, EPO converts the predictive density into the probability that
the realized outcome lies inside the task tolerance box around the planner
target. For task-relevant coordinates $\mathcal I$ with tolerances
$\epsilon_d$, the coordinate-wise probability is
\[
\hat C_{k,d}(s,z;z^\star)
=
\Phi\!\left(
\frac{z^\star_d+\epsilon_d-\mu_{k,d}(s,z)}{\sigma_{k,d}(s,z)}
\right)
-
\Phi\!\left(
\frac{z^\star_d-\epsilon_d-\mu_{k,d}(s,z)}{\sigma_{k,d}(s,z)}
\right),
\]
with $\Phi$ the standard Gaussian \rev{CDF and $(\mu_{k,d}-z^\star_d)$ wrapped
to $(-\pi,\pi]$ for orientation coordinates. In our experiments
$\mathcal I$ contains all 16 coordinates in both modes. The consistency
probability is} the product over \rev{these} coordinates,
$\hat C_k(s,z;z^\star)=\prod_{d\in\mathcal I}\hat C_{k,d}(s,z;z^\star)$.
For numerical stability, we lower-bound each coordinate probability as
$\tilde C_{k,d}=\max\{\hat C_{k,d},10^{-8}\}$ and the feasibility
probability as $\tilde F_k=\max\{\hat F_k,10^{-3}\}$. The runtime energies
are $\hat E_k^{\mathrm C}=-\sum_{d\in\mathcal I}\log\tilde C_{k,d}$ and
$\hat E_k^{\mathrm F}=-\log\tilde F_k$, giving the full learned energy
\[
\hat E_k(s,z;z^\star)
=
\hat E_k^{\mathrm F}(s,z)+\hat E_k^{\mathrm C}(s,z;z^\star).
\]
Both terms are probability-derived negative-log energies, so we do not introduce an additional hand-tuned weight between feasibility and consistency.

\subsection{Runtime optimizer}
\label{app:impl-runtime}

EPO refines goals at $10$~Hz and re-scores candidates and selects policies
at the $50$~Hz control rate. At each refinement update (every $T_r=5$
control ticks), the planner target $z^\star$ is projected into each
task-compatible policy's goal subspace, $z_k^{(0)}=\rho_k(z^\star)$.
Refinement is applied only when $\hat F_k(s,z_k^{(0)})\ge\tau$, with
$\tau=0.9$. The operator $\operatorname{Refine}_{n,\eta}$ in
Algorithm~\ref{alg:tto} takes $n=3$ projected gradient steps on
$\hat E_k(s,z;z^\star)$ with step size $\eta=0.05$:
\[
z_k^{(i+1)}
=
\mathrm{clip}_{\mathcal Z_k}
\left[z_k^{(i)}-\eta\frac{g_k^{(i)}}{\max\{1,\|g_k^{(i)}\|_2\}}\right],
\qquad
g_k^{(i)}=M_k\nabla_z\hat E_k(s,z_k^{(i)};z^\star).
\]
Here $M_k$ retains the active base-velocity and torso-height coordinates;
end-effector targets remain fixed. The gradient norm is clipped at $1$,
and the clipping operator projects the command back into the valid goal
box $\mathcal Z_k$. We use no additional trust-region projection beyond
the policy-specific command bounds. \rev{Refinement returns the lowest-energy
point among $z_k^{(0)},\ldots,z_k^{(n)}$, which need not be the last
iterate.} The resulting goals are cached until the next refinement update.
\rev{A standing policy is task-compatible only when the commanded base velocity
lies within tolerance of zero ($|v_x|,|v_y|\le0.20$~m/s,
$|\omega_z|\le0.50$~rad/s); its velocity} command is \rev{then fixed to zero.
Removing this rule changes the EPO-M4 results only marginally
(loco-manipulation error $0.546$ in both cases, push survival $95.8\%$
versus $95.3\%$; standing manipulation unchanged at $0.420$ and $97.7\%$;
$2.21$ versus $2.08$ switches per episode).}

At every control tick, EPO re-evaluates the cached goals using the latest
state and applies the stateful operator
$\operatorname{Select}_{\tau,B,N_e}$ in Algorithm~\ref{alg:tto}. Write
$F_k=\hat F_k(s_t,\hat z_{k,t})$,
$e_k=\hat E_k(s_t,\hat z_{k,t};z_t^\star)$, and let $i=k_{t-1}$ be the
current policy. While $F_i\ge\tau$, each task-compatible challenger with
$F_j\ge\tau$ accumulates its energy advantage:
\[
A_j\leftarrow\max\{0,A_j+e_i-e_j\}.
\]
Challengers below $\tau$ have $A_j=0$. EPO switches to
$\arg\max_j A_j$ when $\max_j A_j>B$, with $B=40$. If instead the
current policy remains below $\tau$ for $N_e=3$ consecutive ticks, EPO
selects the lowest-energy candidate meeting $\tau$, or the candidate
with the highest predicted feasibility if none qualifies. Accumulated
advantages are cleared after switching. The selected command is sent to
the corresponding policy. \rev{Since the selector runs at $50$~Hz, $B$ is
accumulated over $50$~Hz samples and $N_e=3$ corresponds to $60$~ms.
Evaluation is synchronous: refinement, scoring, and selection finish
within the control tick before the command is applied.}
\subsection{Policy family training}
\label{app:impl-policy}

We train four goal-conditioned whole-body controllers
$\Pi=\{\pi_{\mathrm{gen}},\pi_{\mathrm{gen}}^R,\pi_{\mathrm{stand}},\pi_{\mathrm{stand}}^R\}$ in
IsaacLab with PPO on rsl-rl. All four share the same network
architecture, observation interface, action space, terrain distribution,
reward design, and PPO optimizer; they differ only in (i) command ranges,
(ii) the magnitude of external disturbances, and (iii) wrist payloads.
Each policy is trained from scratch in simulation, with no fine-tuning,
distillation, or real-world adaptation.

\paragraph{Robot and control.}
We simulate the 29-DoF Unitree G1. Joint targets are tracked by IsaacLab's
implicit PD actuator, with stiffness, damping, and armature from the
official Unitree URDF (representative gains $k_p=200$ on hip-pitch and
knee, $20$ on ankles, $100$ on shoulders, $40$ on wrists; damping
proportional to stiffness). Physics is integrated at $200$~Hz with control
decimation $4$, yielding a $50$~Hz policy rate.

\paragraph{Architecture.}
Actor and critic are recurrent: a single-layer LSTM (hidden size $256$)
followed by a three-layer MLP of widths $[256,256,128]$ with ELU. The
actor parameterizes a diagonal Gaussian with a learned scalar
log-standard-deviation ($\sigma_0=1.0$). The critic consumes privileged
observations (body-frame linear velocity, binary foot-contact
indicators, and gait-clock phase) and uses rsl-rl empirical normalization;
the deployed actor uses raw observations both in simulation and on hardware.
Gait-clock phase is provided only to the critic.

\paragraph{Observation and action spaces.}
At each step the actor concatenates a proprioceptive block (root angular
velocity, projected gravity, joint position and velocity relative to the
nominal pose, and previous action), the 16D goal command
$g=[v_x,v_y,\omega_z,p^\ell_{ee},p^r_{ee},\theta^\ell_{ee},\theta^r_{ee},h]$,
and a terrain height scan ($187$ ray casts on a $17\times11$ grid,
$0.1$~m spacing). Inactive goal dimensions are zeroed throughout training
and deployment, so all specialists share one interface. The actor outputs
a $29$-dim action scaled by $0.25$ to joint-position targets and clipped
to $90\%$ of each joint limit. Observation noise is injected during training;
the deployed actor receives the same interface from on-board estimators.

\paragraph{PPO.}
PPO collects $24$ steps from each of $N_{\rm env}=8192$ environments per
iteration, with $5$ epochs over $4$ minibatches (clip $0.2$, value
coefficient $1.0$, entropy coefficient $0.005$, $\gamma=0.99$,
$\lambda=0.95$, gradient-norm cap $1.0$). The learning rate uses rsl-rl's
adaptive schedule (target KL $0.01$, initialized at $1\times10^{-3}$).
Each policy is trained for $2\times10^4$ iterations on a single
NVIDIA RTX 5090, with random seed $42$ unless noted.

\paragraph{Rewards.}
The per-step reward is a weighted sum of task-tracking, gait, and
regularization terms and is identical for all four policies. All policies
track bilateral end-effector pose, torso height, planar base velocity,
and yaw rate through an exponential kernel
$\rho(e;\sigma)=\exp(-\|e\|^2/\sigma^2)$; the standing specialists use zero
base-velocity commands. End-effector tracking is additionally rewarded in
joint space using inverse kinematics. Gait is shaped by a phase clock
with a $0.64$~s cycle and phase locking to measured contacts: rewards
track a reference joint trajectory, $6$~cm swing-foot clearance, and the
clock's contact schedule, with additional foot-contact-force and low-speed
terms. Regularization penalizes torso tilt, joint-configuration deviation,
action rate, joint acceleration, arm joint velocity and acceleration,
vertical and angular base velocity, mechanical power, undesired or lost
contacts, foot sliding, and joint-limit violations.
Contact parameters, inertia, and the initial state are randomized per episode
($\mu_s\sim U(0.6,1.0)$, $\mu_d\sim U(0.4,0.8)$,
$\Delta m_{\rm torso}\sim U(-5,5)$~kg; root pose, velocities, and joint
configuration perturbed at reset).

\paragraph{Per-policy differences.}
$\pi_{\mathrm{gen}}$ and $\pi_{\mathrm{gen}}^R$ activate all 16 goal
dimensions, with $v_x\in[-0.6,1.0]$~m/s, $v_y\in[-0.5,0.5]$~m/s,
and $\omega_z\in[-1.57,1.57]$~rad/s resampled every $10$~s; $20\%$ of
environments receive standing commands. The two standing specialists
train under zero base velocity: the velocity dimensions are masked and
the active goal subspace reduces to
$\{p^{\ell/r}_{ee},\theta^{\ell/r}_{ee},h\}$.
For all policies, end-effector positions are sampled in the body frame
with $x\in[0.08,0.37]$~m, $|y|\in[0.15,0.37]$~m, and
$z\in[-0.04,0.40]$~m, and torso height lies in
$[0.6h_0,1.2h_0]$, where $h_0$ is the nominal height.
End-effector and torso-height targets resample every $5$--$10$~s;
the standing specialists use instantaneous end-effector resampling.
$\pi_{\mathrm{gen}}$ and $\pi_{\mathrm{stand}}$ receive base-velocity
perturbations up to $\pm1$~m/s every $10$--$15$~s.
The two robust variants, $\pi_{\mathrm{gen}}^R$ and
$\pi_{\mathrm{stand}}^R$, use perturbations up to $\pm2$~m/s with
per-wrist payloads $\sim U(0,2)$~kg added at reset.
The four specialists thus form a compact family with complementary
strengths in locomotion and standing manipulation under nominal and
disturbed conditions.
\subsection{Baselines and runtime variants}
\label{app:impl-baselines}

Within each policy family, all methods share the frozen low-level policies,
the planner target $z^\star$, the same physics, observations, and hardware
safety limits. Energy-based variants use the models of
\S\ref{app:impl-energy}. The only differences are which
policy executes at each step and whether the executed command is the
projected target $z^\star$ or the refined command from the inner-loop
optimizer.

\paragraph{Fixed-policy baselines.}
Each baseline executes one specialist ($\pi_{\mathrm{gen}}$,
$\pi_{\mathrm{gen}}^R$, or, in standing mode, $\pi_{\mathrm{stand}}$) and
runs the projected target $z^\star$ at every step, without selection or
refinement.

\paragraph{Single-policy goal optimization (+Opt).}
For a fixed policy, +Opt applies the same inner-loop gradient refinement
that EPO uses ($n=3$ steps, $\eta=0.05$), but does not switch policies.
This isolates the value of command refinement with the cross-policy
selector disabled.

\paragraph{Learned MoE scheduler.}
\rev{This baseline replaces EPO's energy-based selector with a learned
categorical actor that picks one specialist every $5$ control ticks
($10$~Hz). Actor and critic are MLPs with hidden widths $(256,128)$,
LayerNorm, and ELU; the input is the $67$-D proprioceptive state, the
$187$-D height scan, and the $16$-D operator command. It is trained with
PPO on top of the frozen policy library ($256$ environments, $200$
iterations of $96$ decisions, $5$ epochs $\times$ $4$ minibatches,
learning rate $3\times10^{-4}$, clip $0.2$, $\gamma=0.99$, $\lambda=0.95$,
entropy coefficient $0.01$, value coefficient $0.5$) on mixed terrain with
random pushes of up to $3$~m/s per axis every $10$--$15$~s and $35\%$
standing environments. The reward per decision is
$-\min(e,100)+0.5\,\mathbf 1[\text{upright}]-100\,\mathbf 1[\text{fall}]$,
where $e$ is the mean normalized squared tracking error over the 16
command dimensions; it contains no energy term. We use the final
checkpoint and take the highest-scoring policy at test time. We report variants on the two-, three-, and
four-policy libraries (MoE-M2, MoE-M3, and MoE-M4). The scheduler switches
among policies but does not refine the continuous command, providing a
comparison to learned discrete selection without command optimization.}

\paragraph{\rev{Training cost and memory.}}
\rev{EPO trains the energy models once per policy, which takes about $30$
GPU-minutes and adds $3.55$~MB per policy. The MoE scheduler takes about
$40$ and $60$ GPU-minutes for $K=2$ and $K=3$ and must be retrained
whenever the policy library changes, whereas adding a policy to EPO
leaves the existing energy models unchanged. Masking a trained gate is
not a substitute for retraining: on identical observations, a $K=3$ gate
masked to two experts disagrees with a gate trained on those two experts
on $4.6$--$16.7\%$ of decisions (range over the three pairs). Covering
every subset of $K$ policies would require up to $2^K-1$ gate variants,
whereas EPO's per-policy energy models apply to any subset.}

\paragraph{EPO variants and ablations.}
\rev{EPO-M2 uses the base family, EPO-M3 adds $\pi_{\mathrm{stand}}$, and
EPO-M4 further adds $\pi_{\mathrm{stand}}^R$; all three share policy and
energy-model checkpoints, runtime parameters, and command sequences. Each
ablation \rev{modifies} Full EPO \rev{as follows}. \emph{No goal refinement} sets
$n=0$. \emph{No refinement gate} refines regardless of $\hat F$.
\emph{No feasibility threshold} sets $\tau=0$ in the selector while
keeping the refinement gate. \emph{No switching threshold} sets $B=0$.
\emph{Feasibility only} and \emph{Consistency only} change only the
selection score to $\log(\hat F^{\alpha}\hat C)$ with $\alpha=100$ and
$\alpha=10^{-6}$, respectively; the refinement objective is unchanged, and
\emph{Consistency only} additionally removes $\tau$ and the refinement
gate.}

\subsection{Simulation evaluation protocol}
\label{app:impl-sim}

\paragraph{Modes and terrains.}
We evaluate in two modes: \emph{loco-manipulation} (base velocity, yaw
rate, torso height, end-effector pose) and \emph{standing manipulation}
(velocity inactive; torso height and end-effector pose). Each mode is
evaluated on three terrains $\{$\rev{flat, rough, stair}$\}$, and all reported
results are averaged over terrains. Each setting uses $128$ matched seeds
per terrain, giving $384$ episodes.

\paragraph{Stress conditions.}
\rev{Each episode lasts $20$~s ($1000$ control ticks). \emph{Normal} resamples
the full command from the training range every $4$~s
($v_x\in[-0.6,-0.1]\cup[0.1,1.0]$~m/s, $v_y\in[-0.5,0.5]$~m/s,
$\omega_z\in[-1.57,1.57]$~rad/s; zero velocity in standing manipulation)
without disturbance and is used to measure tracking precision.
\emph{Extreme} alternates every $2$~s between two opposite corners of the
command range, testing stability at the boundary of the family's operating
region. \emph{Push} holds a fixed carry target (hands at
$(0.27,\pm0.20,0.10)$~m in the base frame, height ratio $0.74$, walking at
$v_x=0.5$~m/s in loco-manipulation) and applies one horizontal $3$~m/s
velocity impulse in a uniformly random direction at a uniformly random time
between $10$ and $15$~s. Environment pushes, reset noise, and observation
noise are disabled.
All methods are evaluated under identical command streams, terrain
instances, and disturbance schedules, so differences reflect the selection
and refinement mechanisms rather than the evaluation distribution.}

\paragraph{Metrics.}
\rev{Survival is the percentage of episodes without a failure, defined as
torso contact, base tilt beyond about $60^\circ$, or base height below
$0.3$~m. Tracking error is the normalized absolute error
$|y_{t,d}-z^\star_{t,d}|/\epsilon_d$ against the operator command,
pooled over the ticks before failure and over the command dimensions $A$
of the mode ($16$ in loco-manipulation, $13$ in standing manipulation):
the total error in a cell is divided by the total number of
dimension--ticks. Tolerances are $\epsilon=0.20$~m/s for $v_x,v_y$,
$0.50$~rad/s for $\omega_z$, $0.05$~m for end-effector position,
$0.10$~rad for each component of end-effector orientation (a base-frame
rotation vector; differences are taken per component and wrapped to
$(-\pi,\pi]$), and $0.10$ for torso height expressed as a ratio to the
height at reset. Values below $1.0$ indicate average tracking within
tolerance. In Push episodes the $4$~s after the push are excluded from the
error.
The precision axis of the main-text Pareto plot is the Normal-condition error averaged over terrains; the robustness axis is a weighted survival score that combines the Normal, Extreme, and Push conditions (each averaged over terrains) under equal weights $w_c = 1/3$ fixed before evaluation. Because Normal-condition survival is near-saturated across methods, the weighted score is in practice dominated by the Extreme and Push conditions.
Switch counts are averaged over Normal and Push episodes. Paired
comparisons match episodes by terrain, seed, and condition and resample
whole seeds ($10{,}000$ resamples, $95\%$ percentile intervals).}

\paragraph{Aggregate results.}
Table~\ref{tab:app-aggregate} reports the aggregate precision--robustness
quantities behind the main-text Pareto analysis, per evaluation mode and
method. Error is the Normal-condition tracking error and Survival is the
weighted survival score described above. EPO reaches low error at high
survival in both modes. Relative to MoE-M4, EPO-M4 reduces tracking error
in both modes, with a larger improvement in standing manipulation.
Adding the standing specialists lowers standing error from $0.456$
(EPO-M2) to $0.422$ (EPO-M3) and $0.420$ (EPO-M4).
EPO-M2, M3, and M4 make $1.85$, $1.91$, and $2.08$ switches per episode,
respectively, compared with $7.35$, $14.56$, and $15.46$ for the matching
MoE variants.

\begin{table}[t]
\centering
\caption{\textbf{Aggregate precision--robustness results.} Error is the
Normal-condition tracking error averaged over (\rev{flat, rough, stair});
Survival is the weighted survival score over the Normal, Extreme, and
Push conditions, each averaged over the same terrains. Lower error and
higher survival are better. These are the aggregate quantities plotted in
Fig.~\ref{fig:pareto}. Best EPO entries in bold.}
\label{tab:app-aggregate}
\setlength{\tabcolsep}{8pt}\footnotesize
\begin{tabular}{lcccc}
\toprule
& \multicolumn{2}{c}{Loco-manip} & \multicolumn{2}{c}{Stand} \\
\cmidrule(lr){2-3}\cmidrule(lr){4-5}
Method & Error $\downarrow$ & Surv (\%) $\uparrow$ & Error $\downarrow$ & Surv (\%) $\uparrow$ \\
\midrule
$\pi_{\mathrm{stand}}$     & \rev{0.656} & \rev{77.0} & 0.396 & 76.0 \\
$\pi_{\mathrm{gen}}$       & 0.549 & 93.6 & 0.458 & 88.3 \\
$\pi_{\mathrm{gen}}^R$     & 0.568 & 97.7 & 0.469 & 98.2 \\
\midrule
$\pi_{\mathrm{stand}}$+Opt & \rev{0.668} & \rev{76.6} & 0.400 & 76.0 \\
$\pi_{\mathrm{gen}}$+Opt   & 0.539 & 93.1 & 0.457 & 87.3 \\
$\pi_{\mathrm{gen}}^R$+Opt & 0.557 & 97.5 & 0.465 & 97.7 \\
\midrule
MoE-M2                    & 0.565 & 97.7 & 0.470 & 98.3 \\
MoE-M3                    & 0.559 & 97.4 & 0.459 & 97.4 \\
MoE-M4                    & 0.568 & 97.7 & 0.461 & 98.3 \\
\midrule
\textbf{EPO-M2}           & \textbf{0.542} & \textbf{96.8} & 0.456 & 97.4 \\
\textbf{EPO-M3}           & 0.546 & 96.5 & 0.422 & 96.3 \\
\textbf{EPO-M4}           & 0.546 & 96.5 & \textbf{0.420} & \textbf{97.7} \\
\bottomrule
\end{tabular}
\end{table}

In the paired Normal and Push evaluation, the EPO-M4 minus MoE-M4
difference in mean clean tracking error is $-0.0306$ ($95\%$ CI:
$[-0.0348,-0.0266]$). Across the $768$ Push episodes, the corresponding
difference in fall rate is $+1.82$ percentage points ($95\%$ CI:
$[0.39,3.39]$).

\subsection{\rev{Recovery from base-policy failures}}
\label{app:impl-recovery}
\begingroup\revcolor
Table~\ref{tab:app-recovery} restricts the Push evaluation to the
episodes in which a fixed policy falls and reports how often EPO survives
the same push (identical seeds, terrains, and push times; loco and
standing pooled).

\begin{table}[h]
\centering
\caption{\textbf{Survival of EPO on episodes where a fixed policy falls}
(Push condition). $n$ is the number of such episodes.}
\label{tab:app-recovery}
\setlength{\tabcolsep}{8pt}\footnotesize
\begin{tabular}{lccc}
\toprule
Fixed policy that falls & $n$ & EPO-M2 survives (\%) & EPO-M4 survives (\%) \\
\midrule
$\pi_{\mathrm{stand}}$ & 503 & 94.6 & 96.8 \\
$\pi_{\mathrm{gen}}$ & 197 & 92.4 & 93.4 \\
$\pi_{\mathrm{gen}}^R$ & 14 & 57.1 & 78.6 \\
\bottomrule
\end{tabular}
\end{table}
\endgroup

\subsection{Runtime and parameter sensitivity}
\label{app:impl-runtime-sensitivity}

\rev{Table~\ref{tab:app-runtime} reports per-robot orchestration latency on a
single NVIDIA RTX PRO 6000 Blackwell. The timed region covers refining and
scoring every policy in the library for one robot, with GPU
synchronization before and after; it excludes policy inference,
observation processing, the selector update, and the $50$~Hz re-scoring
between refinements. Each configuration uses $150$ measurements after
discarding $50$ warm-up samples; Max Hz is the inverse median latency.
The ablation table in the main text uses a separate measurement per
variant (e.g.\ $615$~Hz instead of $601$~Hz for EPO-M2 without
refinement). The full refinement cycle remains faster than the
deployed $10$~Hz refinement rate for all three policy families.}

\begin{table}[t]
\centering
\caption{\textbf{Runtime scaling.} \rev{Latency of refining and scoring all
policies for one robot,} with three refinement steps and without
refinement.}
\label{tab:app-runtime}
\setlength{\tabcolsep}{5pt}\footnotesize
\begin{tabular}{lcccccc}
\toprule
& \multicolumn{3}{c}{Three refinement steps}
& \multicolumn{3}{c}{No refinement} \\
\cmidrule(lr){2-4}\cmidrule(lr){5-7}
Family & Median (ms) & p95 (ms) & Max Hz
& Median (ms) & p95 (ms) & Max Hz \\
\midrule
EPO-M2 & 12.06 & 12.48 & 83 & 1.67 & 1.80 & 601 \\
EPO-M3 & 20.46 & 20.81 & 49 & 2.40 & 2.55 & 416 \\
EPO-M4 & 24.34 & 27.48 & 41 & 3.18 & 3.34 & 314 \\
\bottomrule
\end{tabular}
\end{table}

\rev{For EPO-M4, we vary one selector parameter at a time from the default
$(\tau,B,N_e)=(0.9,40,3)$, testing $\tau\in\{0.80,0.95\}$,
$B\in\{20,80\}$, and $N_e=5$, with $128$ matched seeds per terrain.
Across these settings, the absolute change in mean clean tracking error
is at most $0.0007$, and the paired $95\%$ confidence intervals for
changes in push fall rate all include zero ($N_e=5$:
$+1.56$~pp, $[0.00,3.26]$). Raising $\tau$ to $0.95$ increases switching
to $2.41$ times per episode. Lowering $B$ to $20$
increases switching to $2.91$ times per episode; raising it to $80$
reduces switching to $1.57$, compared with $2.08$ at the default.}

\begingroup\revcolor\subsection{Horizon sensitivity}
\label{app:impl-horizon}\endgroup

\rev{Table~\ref{tab:app-horizon} varies the rollout horizon $H$ used to train
the energy models of EPO-M4. The three retrained rows use an identical
training recipe and differ only in $H$; the selector and refinement
parameters are those tuned for $H=50$. Each row uses $128$ seeds on each
of three terrains. The models used in the main experiments, from a separate
training run at $H=50$, are shown for reference.}

\begin{table}[t]
\centering\revcolor
\caption{\textbf{Energy-model horizon.} EPO-M4 with energy models trained
at different horizons. Err.\ is Normal tracking error; Surv.\ is survival
under a $3$~m/s push; Push fall pools both modes. $^\dagger$: \rev{push fall rate} differs from
the retrained $H=50$ (paired whole-seed bootstrap, $95\%$ CI excludes 0).}
\label{tab:app-horizon}
\setlength{\tabcolsep}{4pt}\footnotesize
\resizebox{\linewidth}{!}{%
\begin{tabular}{lcccccc}
\toprule
$H$ (control ticks) & Loco Err $\downarrow$ & Stand Err $\downarrow$
& Loco Surv.\ $\uparrow$ & Stand Surv.\ $\uparrow$
& \rev{Push fall (\%)} $\downarrow$ & \rev{Switches/ep} \\
\midrule
$10$ ($0.2$~s)  & 0.547 & 0.433 & 97.9 & 80.5 & 10.8$^\dagger$ & 1.49 \\
$50$ ($1.0$~s)  & 0.550 & 0.428 & 95.6 & 98.4 & 3.0 & 1.60 \\
$100$ ($2.0$~s) & 0.554 & 0.429 & 97.4 & 96.9 & 2.9 & 1.40 \\
\midrule
$50$, main-table models & 0.546 & 0.420 & 95.3 & 97.7 & 3.5 & 2.08 \\
\bottomrule
\end{tabular}}
\end{table}

\rev{A short horizon loses robustness: with $H=10$ the push fall rate rises
from $3.0\%$ to $10.8\%$ ($+7.8$~pp, $95\%$ CI $[4.8,11.1]$), entirely
in standing manipulation, while Normal tracking error does not change
significantly ($+0.0007$, $[-0.0027,0.0042]$). A longer horizon brings no
benefit: $H=100$ leaves the fall rate unchanged ($-0.13$~pp,
$[-1.69,1.56]$) and slightly increases Normal error ($+0.0032$,
$[0.0006,0.0058]$). We therefore use $H=50$, one second at the $50$~Hz
control rate.}

\end{document}